\documentclass[11pt]{article}
\usepackage{colacl}
\usepackage{graphicx}
\usepackage{psfrag}

\title{Prefix Probabilities from Stochastic Tree Adjoining
  Grammars\thanks{Part of this research was done while the first and
    the third authors were visiting the Institute for Research in
    Cognitive Science, University of Pennsylvania.  The first author
    was supported by the German Federal Ministry of Education,
    Science, Research and Technology (BMBF) in the framework of the
    {\sc Verbmobil} Project under Grant 01 IV 701 V0, and by the
    Priority Programme Language and Speech Technology, which is
    sponsored by NWO (Dutch Organization for Scientific Research).
    The second and third authors were partially supported by NSF grant
    SBR8920230 and ARO grant DAAH0404-94-G-0426.  The authors wish to
    thank Aravind Joshi for his support in this research.}}

\author{Mark-Jan Nederhof\\
DFKI\\
Stuhlsatzenhausweg 3,\\
         D-66123 Saarbr{\"u}cken, \\
         Germany \\
{\tt nederhof@dfki.de}
\And
Anoop Sarkar \\
Dept. of Computer and Info. Sc. \\
Univ of Pennsylvania \\
200 South 33rd Street, \\
Philadelphia, PA 19104 USA \\
{\tt anoop@linc.cis.upenn.edu}
\And
Giorgio Satta \\
Dip. di Elettr. e Inf. \\
Univ. di Padova \\
via Gradenigo 6/A, \\
35131 Padova, Italy \\
{\tt satta@dei.unipd.it}}

\newcommand{\deltafunc}{\delta}

\newcommand{\mycomment}[1]{}

\newcommand{\nonterm}{{\it NT}}
\newcommand{\myterm}{\Sigma}
\newcommand{\myvarbot}{V^\bot}
\newcommand{\domfoot}{{\it dft}}
\newcommand{\myspan}{\sigma}
\newcommand{\myvar}{V}

\newcommand{\order}[1]{{\cal O}(#1)}

\newcommand{\longsum}[1]{\sum_{\makebox[5ex]{\scriptsize$#1$}}}

\newcommand{\Plowest}{P_{\scriptscriptstyle low}}
\newcommand{\Phighest}{P_{\scriptscriptstyle split}}
\newcommand{\Poutside}{P_{\scriptscriptstyle outer}}

\newcommand{\ep}{\epsilon}

\newcommand{\lab}{\mbox{{\it label\/}}}
\newcommand{\children}{\mbox{{\it cdn\/}}}

\begin{document}

\maketitle

\begin{abstract}
  Language models for speech recognition typically use a probability
  model of the form $\Pr( a_n | a_1, a_2, \ldots, a_{n-1})$.
  Stochastic grammars, on the other hand, are typically used to assign
  structure to utterances. A language model of the above form is
  constructed from such grammars by computing the prefix probability
  $\sum_{w \in \myterm^\ast} \Pr(a_1 \cdots a_n w)$, where $w$
  represents all possible terminations of the prefix $a_1 \cdots a_n$.
  The main result in this paper is an algorithm to compute such prefix
  probabilities given a stochastic Tree Adjoining Grammar (TAG). The
  algorithm achieves the required computation in ${\cal O}(n^6)$ time.
  The probability of subderivations that do not derive any words in
  the prefix, but contribute structurally to its derivation, are
  precomputed to achieve termination. This algorithm enables existing
  corpus-based estimation techniques for stochastic TAGs to be used
  for language modelling.
\end{abstract}

\section{Introduction}
\label{intro}

Given some word sequence $a_1 \cdots a_{n-1}$, speech recognition
language models are used to hypothesize the next word $a_n$, which
could be any word from the vocabulary $\myterm$. This is typically
done using a probability model $\Pr( a_n | a_1, \ldots, a_{n-1}
)$. Based on the assumption that modelling the hidden structure of
natural language would improve performance of such language models,
some researchers tried to use stochastic context-free grammars (CFGs)
to produce language models
\cite{Wright.Wrigley,Jelinek.Lafferty,Stolcke}.  The probability model
used for a stochastic grammar was $\sum_{w \in \myterm^\ast} \Pr(a_1
\cdots a_n w)$.  However, language models that are based on trigram
probability models out-perform stochastic CFGs. The common wisdom
about this failure of CFGs is that trigram models are lexicalized
models while CFGs are not.

Tree Adjoining Grammars (TAGs) are important in this respect since
they are easily lexicalized while capturing the constituent structure
of language.  More importantly, TAGs allow greater linguistic
expressiveness. The trees associated with words can be used to encode
argument and adjunct relations in various syntactic environments. This
paper assumes some familiarity with the TAG formalism.
\cite{Joshi.MOLbook} and \cite{Joshi.Schabes.TreeAutomatabook} are
good introductions to the formalism and its linguistic relevance.
TAGs have been shown to have relations with both phrase-structure
grammars and dependency grammars \cite{Rambow.Joshi.MTTbook}, which is
relevant because recent work on {\em structured} language models
\cite{Hopkins.summer} have used dependency grammars to exploit their
lexicalization. We use stochastic TAGs as such a {\em structured}
language model in contrast with earlier work where TAGs have been
exploited in a class-based $n$-gram language model~\cite{Srini.ICSLP}.

This paper derives an algorithm to compute prefix probabilities 
$\sum_{w \in \myterm^\ast} \Pr(a_1 \cdots a_n w)$.
The algorithm assumes as input a stochastic TAG $G$ and a string which
is a prefix of some string in
$L(G)$, the language generated by $G$. This algorithm enables existing
corpus-based estimation techniques \cite{Schabes.Coling92} in
stochastic TAGs to be used for language modelling.

\section{Notation}
\label{notation}

A stochastic Tree Adjoining Grammar (STAG)
is represented by a tuple $(\nonterm, \myterm, {\cal
  I}, {\cal A}, \phi)$ where $\nonterm$ is a set of
nonterminal symbols,
$\myterm$ is a set of terminal symbols,
${\cal I}$ is a set of {\bf initial} trees and
${\cal A}$ is a set of {\bf auxiliary} trees.
Trees in ${\cal I} \cup {\cal A}$ are also called {\bf elementary} trees.

We refer to the root
of an elementary tree $t$ as $R_t$. Each auxiliary tree has exactly one
distinguished leaf, which is called the {\bf foot}.
We refer to the foot of an
auxiliary tree $t$ as $F_t$.
We let $\myvar$ denote the set of all nodes in the elementary trees.

For each leaf $N$ in an elementary tree, except when it is a foot, we define
$\lab(N)$ to be the label of the node, which is either
a terminal from $\myterm$ or the empty string $\epsilon$.
For each other node $N$, $\lab(N)$ is an element from $\nonterm$.

At a node $N$ in a tree such that $\lab(N)\in \nonterm$
an operation called {\bf adjunction\/}
can be applied, which excises the tree at $N$ and inserts an auxiliary tree.

Function $\phi$ assigns a probability to each adjunction.
The probability of adjunction of $t\in {\cal A}$ at node $N$ is denoted by
$\phi(t,N)$. The probability that at $N$ no adjunction is
applied is denoted by $\phi({\bf nil},N)$.
We assume that each STAG $G$ that we consider is
{\bf proper}. That is, for each $N$ such that $\lab(N)\in \nonterm$,
$${\sum_{t\in {\cal A}\cup\{{\bf nil\}}}} \phi(t,N) =  1.$$

For each non-leaf node
$N$ we construct the string $\children(N)=\widehat{N_1}\cdots\widehat{N_m}$
from the (ordered) list of
children nodes $N_1,\ldots,N_m$ by defining, for each $d$
such that $1\leq d\leq m$, $\widehat{N_d}= \lab(N_d)$
in case $\lab(N_d) \in \myterm\cup\{\ep\}$, and
$\widehat{N_d}=N_d$ otherwise. In other words, children nodes are replaced by 
their labels unless the labels are nonterminal symbols.

To simplify the exposition, we assume an additional node for each
auxiliary tree $t$, which we denote by $\bot$. This is the unique
child of the actual foot node $F_t$. That is, we change the definition
of $\children$ such that $\children(F_t)=\bot$ for each auxiliary tree
$t$. We set
$$\myvarbot= \{N\in \myvar\ |\ \lab(N)\in \nonterm\} \cup \myterm
\cup \{\bot\}.$$

We use
symbols $a, b, c, \ldots$ to range over $\myterm$,
symbols $v, w, x, \ldots$ to range over $\myterm^\ast$,
symbols $N, M, \ldots$ to range over $\myvarbot$,
and symbols $\alpha, \beta, \gamma, \ldots$ to range over $(\myvarbot)^\ast$.
We use $t, t', \ldots$ to denote trees in
${\cal I} \cup {\cal A}$ or subtrees thereof.

We define the predicate $\domfoot$ on elements from
$\myvarbot$ as $\domfoot(N)$ if and only if (i)~$N\in V$ and $N$
dominates $\bot$, or (ii)~$N=\bot$.
We extend $\domfoot$ to strings of the form
$N_1\ldots N_m\in (\myvarbot)^\ast$ by defining
$\domfoot(N_1\ldots N_m)$ if and only if
there is a $d$ ($1\leq d\leq m$) such that $\domfoot(N_d)$.

For some logical expression $p$, we define
$\deltafunc(p) = 1$ iff $p$ is true, $\deltafunc(p) = 0$ otherwise.  

\section{Overview}
\label{s:overview}

The approach we adopt in the next section to derive a method for the
computation of prefix probabilities for TAGs is based on
transformations of equations.  Here we informally discuss the general
ideas underlying equation transformations. 

Let $w = a_1 a_2 \cdots a_n\in \myterm^*$ be a string
and let $N\in \myvarbot$.
We use the following representation which is standard in tabular methods for
TAG parsing.
An {\bf item} is a tuple 
$[N, i,j, f_1,f_2]$ representing the set of all
trees $t$ such that (i)~$t$ is a subtree rooted at $N$ of some derived
elementary tree; and (ii)~$t$'s root spans from position $i$ to
position $j$ in $w$, $t$'s foot node spans from position $f_1$ to
position $f_2$ in $w$.  In case $N$ does not dominate the foot, we set
$f_1 = f_2 = -$.  
We generalize in the obvious way to items $[t, i,j, f_1,f_2]$,
where $t$ is an elementary tree, and
$[\alpha, i,j, f_1,f_2]$, where $\children(N)=\alpha\beta$ for
some $N$ and $\beta$.

\begin{figure}[htbp]
  \begin{center}
    \leavevmode
    \psfrag{i}{$\scriptstyle i$}
    \psfrag{j}{$\scriptstyle j$}
    \psfrag{f_1}{$\scriptstyle f_1$}
    \psfrag{f_2}{$\scriptstyle f_2$}
    \psfrag{R}{$\scriptstyle R_t$}
    \psfrag{F}{$\scriptstyle F_t$}
    \psfrag{N}{$\scriptstyle N$}
    \psfrag{R'}{$\scriptstyle R_{t'}$}
    \psfrag{F'}{$\scriptstyle F_{t'}$}
    \psfrag{epsilon}{$\scriptstyle \ep$}
    \psfrag{beta}{$t$}
    \psfrag{beta'}{$t'$}
    \psfrag{spine}{\mbox{\footnotesize spine}}
    \psfrag{\(a\)}{\mbox{(a)}}
    \psfrag{\(b\)}{\mbox{(b)}}
    \includegraphics[height=2.2in]{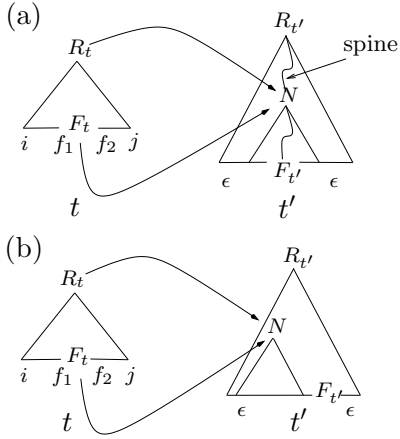}
    \caption{Wrapping in auxiliary trees with empty yield}
    \label{f:wrap}
  \end{center}
\end{figure}

To introduce our approach, let us start with some considerations
concerning the TAG parsing problem.  When parsing $w$ with a TAG $G$,
one usually composes items in order to construct new items spanning a
larger portion of the input string.  Assume there are instances of
auxiliary trees $t$ and $t'$ in $G$, where the yield of $t'$, apart
from its foot, is the empty string.  If $\phi(t,N) > 0$ for some node
$N$ on the spine of $t'$, and we have recognized an item $[R_t, i,j,
f_1,f_2]$, then we may
adjoin $t$ at $N$ and hence deduce the existence of an item $[R_{t'},
i,j, f_1,f_2]$ (see Fig.~\ref{f:wrap}(a)).  Similarly, if $t$ can be
adjoined at a node $N$ to the left of the spine of $t'$ and $f_1=f_2$,
we may deduce the existence of an item $[R_{t'}, i,j, j,j]$ (see
Fig.~\ref{f:wrap}(b)).  Importantly, one or more other auxiliary trees
with empty yield could wrap the tree $t'$ before $t$ adjoins.
Adjunctions in this situation are potentially nonterminating.

\begin{figure}[htbp]
  \begin{center}
    \leavevmode
    \psfrag{i}{$\scriptstyle i$}
    \psfrag{j}{$\scriptstyle j$}
    \psfrag{n}{$\scriptstyle n$}
    \psfrag{f_1}{$\scriptstyle f_1$}
    \psfrag{f_2}{$\scriptstyle f_2$}
    \psfrag{R}{$\scriptstyle R_t$}
    \psfrag{F}{$\scriptstyle F_t$}
    \psfrag{N}{$\scriptstyle N$}
    \psfrag{R'}{$\scriptstyle R_{t'}$}
    \psfrag{F'}{$\scriptstyle F_{t'}$}
    \psfrag{epsilon}{$\scriptstyle \ep$}
    \psfrag{spine}{\mbox{\footnotesize spine}}
    \psfrag{\(a\)}{\mbox{(a)}}
    \psfrag{\(b\)}{\mbox{(b)}}
    \includegraphics[height=1.5in]{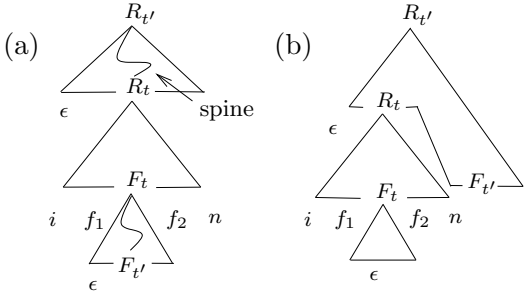}
    \caption{Wrapping of auxiliary trees when computing the prefix probability}
    \label{f:pwrap}
  \end{center}
\end{figure}

One may argue that situations where auxiliary trees have empty yield
do not occur in practice, and are even by definition excluded in the
case of lexicalized TAGs.  However, in the computation of the prefix
probability we must take into account trees with non-empty yield which
behave like trees with empty yield because their lexical nodes fall to
the right of the right boundary of the prefix string.  For example,
the two cases previously considered in Fig.~\ref{f:wrap} now
generalize to those in Fig.~\ref{f:pwrap}.

To derive a method for the computation of prefix probabilities, we
give some simple recursive equations.  Each equation {\em
  decomposes\/} an item into other items in all possible ways, in the
sense that it expresses the probability of that item as a function of
the probabilities of items associated with equal or smaller portions
of the input.

In specifying the equations, we exploit techniques used in the parsing
of incomplete input~\cite{LA88}.  This allows us to compute the prefix
probability as a by-product of computing the inside probability.

In order to avoid the problem of nontermination outlined above, we
transform our equations to remove infinite recursion, while preserving
the correctness of the probability computation.  The transformation of
the equations is explained as follows. For an item $I$, the {\bf span}
of $I$, written $\myspan(I)$, is the $4$-tuple representing the 4
input positions in $I$.  We will define an equivalence relation on
spans that relates to the portion of the input that is covered.  The
transformations that we apply to our equations produce two new sets of
equations. The first set of equations are concerned with all possible
decompositions of a given item $I$ into set of items of which one has
a span equivalent to that of $I$ and the others have an empty span.
Equations in this set represent endless recursion.  The system of all
such equations can be solved independently of the actual input
$w$. This is done once for a given grammar.

The second set of equations have the property that, when evaluated,
recursion always terminates. The evaluation of these equations
computes the probability of the input string
modulo the computation of some parts of the derivation that do not
contribute to the input itself. 
Combination of the second set of equations with the solutions obtained 
from the first set allows the effective
computation of the prefix probability.

\section{Computing Prefix Probabilities}
\label{s:method}

This section develops an algorithm for the computation of 
prefix probabilities for stochastic TAGs.

\subsection{General equations}
\label{ss:general}

The prefix probability is given by:
\begin{eqnarray*}
\label{TAGtop}
\sum_{w \in \myterm^\ast} \Pr(a_1 \cdots a_n w) 
& = & \sum_{t\in {\cal I}} P([t,  0,  n,  -,  -]),
\end{eqnarray*}
where $P$ is a function over items recursively defined as follows:
\begin{eqnarray}
\label{TAGtree}
\lefteqn{P([t , i, j,  f_1,  f_2])\ = P([R_t , i, j,  f_1,  f_2]);} \\
\label{TAGrecur1}
\lefteqn{P([\alpha N , i, j,  -,  -])\ =} \\
&&
\longsum{k(i\leq k \leq j)} P([\alpha , i, k,  -,  -])  \cdot
P([N , k, j,  -,  -]), \nonumber \\
&&\mbox{if\ }
\alpha\neq\ep\wedge \neg\domfoot(\alpha N); \nonumber \\
\label{TAGrecur2}
\lefteqn{P([\alpha N , i, j,  f_1,  f_2])\ =} \\
&&
\longsum{k(i\leq k \leq f_1)} P([\alpha , i, k,  -,  -])  \cdot
P([N , k, j,  f_1,  f_2]), \nonumber\\
&& \mbox{if\ }
\alpha\neq\ep\wedge \domfoot(N);\nonumber  \\
\label{TAGrecur3}
\lefteqn{P([\alpha N , i, j,  f_1,  f_2])\ =} \\
&&
\longsum{k(f_2\leq k \leq j)} P([\alpha , i, k,  f_1,  f_2])  \cdot
P([N , k, j,  -,  -]), \nonumber \\
&&\mbox{if\ }
\alpha\neq\ep\wedge \domfoot(\alpha); \nonumber \\
\label{TAGrule1}
\lefteqn{P([N, i, j, f_1, f_2])\ =} \\
&&
\phi({\bf nil},N)\cdot P([\children(N), i, j, f_1, f_2])\ + \nonumber \\
&& \longsum{\lefteqn{f_1',f_2'(i\leq f_1'\leq f_1 \wedge 
        f_2\leq f_2'\leq j)}\hspace{2cm}}%
P([\children(N), f_1', f_2', f_1,  f_2])\ \cdot\nonumber  \\
&&\hspace{1cm} \sum_{t\in {\cal A}}
     \phi(t,N)\cdot P([t,  i, j, f_1',  f_2']),\nonumber  \\
&& \mbox{if\ } N\in \myvar\wedge\domfoot(N); \nonumber  \\
\label{TAGrule2}
\lefteqn{P([N, i, j, -, -])\ =} \\
&&
\phi({\bf nil},N)\cdot P([\children(N), i, j, -, -])\ + \nonumber \\
&& \longsum{\hspace{1cm}f_1',f_2'(i\leq f_1' \leq f_2'\leq j)}
P([\children(N),  f_1', f_2', -, -])\ \cdot\nonumber  \\
&&\hspace{1cm} \sum_{t\in {\cal A}}
     \phi(t,N)\cdot P([t,  i, j, f_1',  f_2']),\nonumber  \\
&& \mbox{if\ }N\in \myvar\wedge  \neg\domfoot(N); \nonumber  \\
\label{TAGscan}
\lefteqn{P([a , i, j,  -,  -])\ =} \\
&&
\deltafunc(i+1=j\wedge  a_j=a) + \deltafunc(i=j=n); \nonumber \\
\label{TAGfoot}
\lefteqn{P([\bot, i, j,  f_1,  f_2])\ = \deltafunc(i=f_1 \wedge j =f_2);} \\
\label{TAGstop}
\lefteqn{P([\ep, i, j,  -,  -])\ = \deltafunc(i=j).}
\end{eqnarray}
Term $P([t , i, j, f_1, f_2])$ gives the inside probability of all
possible trees derived from elementary tree $t$, having the indicated
span over the input.
This is decomposed into the contribution of each
single node of $t$ in equations~(\ref{TAGtree}) through~(\ref{TAGrule2}).
In equations~(\ref{TAGrule1}) and (\ref{TAGrule2}) the contribution of
a node $N$ is determined by the combination of the inside
probabilities of $N$'s children and by all possible adjunctions at
$N$.
In~(\ref{TAGscan}) we recognize some terminal symbol if it occurs in
the prefix, or ignore its contribution to the span if it occurs after
the last symbol of the prefix.  Crucially, this step allows us to
reduce the computation of prefix probabilities to the computation of
inside probabilities.

\subsection{Terminating equations}
\label{ss:on-line}

In general, the recursive equations~(\ref{TAGtree}) to~(\ref{TAGstop})
are not directly computable.  This is because the value of
$P([A,i,j,f,f'])$ might indirectly depend on itself, giving rise to
nontermination.  We therefore rewrite the equations.

We define an equivalence relation over spans, that expresses when two
items are associated with equivalent portions of the input:

\medskip
\noindent
$\lefteqn{(i',j',f_1',f_2')\approx(i,j,f_1,f_2) 
        \mbox{ if and only if } }$\\[.3em]
\hspace*{.2cm} $((i',j') = (i,j)) \wedge $\\[.3em]
\hspace*{.4cm} $\; \; ((f_1',f_2') = (f_1,f_2) \vee  $\\[.3em]
\hspace*{.4cm} $\; \; ((f_1'=f_2'=i \vee f_1'=f_2'=j 
                \vee f_1'=f_2'=-)\wedge $\\[.3em]
\hspace*{.5cm} $\; \; (f_1=f_2=i\vee f_1=f_2=j \vee f_1=f_2=-)))$ \\[-.2cm]

We introduce two new functions $\Plowest$ and $\Phighest$.  When
evaluated on some item $I$, $\Plowest$ recursively calls itself as
long as some other item $I'$ with a given elementary tree as its first
component can be reached, such that $\myspan(I) \approx \myspan(I')$.
$\Plowest$ returns $0$ if the actual branch of recursion cannot
eventually reach such an item $I'$, thus removing the contribution to
the prefix probability of that branch.  If item $I'$ is reached, then
$\Plowest$ switches to $\Phighest$.  Complementary to $\Plowest$,
function $\Phighest$ tries to decompose an argument item $I$ into
items $I'$ such that $\myspan(I) \not \approx \myspan(I')$.  If this
is not possible through the actual branch of recursion, $\Phighest$
returns $0$. If decomposition is indeed possible, then we start again
with $\Plowest$ at items produced by the decomposition.  The effect of
this intermixing of function calls is the simulation of the original
function $P$, with $\Plowest$ being called only on potentially
nonterminating parts of the computation, and $\Phighest$ being called
on parts that are guaranteed to terminate.

Consider some derivation tree spanning some portion of the input
string, and the associated derivation tree $\tau$.  There must be a
unique elementary tree which is represented by a node in $\tau$ that
is the ``lowest'' one that entirely spans the portion of the input of
interest.  (This node might be the root of $\tau$ itself.)  Then, for
each $t\in{\cal A}$ and for each $i,j,f_1,f_2$ such that $i<j$ and
$i\leq f_1\leq f_2\leq j$, we must have: \vspace{-1ex}
\begin{eqnarray} 
\label{tagnontsum1}
\lefteqn{P([t, i, j, f_1,  f_2])\ =} \\ &&
\longsum{\lefteqn{t'\in{\cal A},f_1',f_2'
((i, j, f_1',  f_2') \approx (i, j, f_1, f_2))}\hspace{3.5cm}}
\Plowest([t, i, j, f_1,  f_2],\ [t',f_1',f_2']). \nonumber \\[-3.0ex] \nonumber
\end{eqnarray}
Similarly, for each $t\in{\cal I}$ and for each $i,j$ such that
$i<j$,  we must have:
\vspace{-1ex}
\begin{eqnarray}
\label{tagnontsum2}
\lefteqn{P([t, i, j, -,  -])\ =} \\ &&
\longsum{t'\in\{t\}\cup{\cal A},f\in\{-,i,j\}}
\Plowest([t, i, j, -,  -],\ [t',f,f]). \nonumber \\[-3.0ex] \nonumber
\end{eqnarray}
The reason why $\Plowest$ keeps a record of indices $f'_1$ and $f'_2$, 
i.e., the spanning of the foot node of the lowest tree 
(in the above sense) on which $\Plowest$ is called, 
will become clear later, when we introduce equations~(\ref{tagoutsideA})
and~(\ref{tagoutsideB}). 

We define $\Plowest([t,i,j,f_1,f_2],[t',f_1',f_2'])$ and
$\Plowest([\alpha ,i,j, f_1, f_2],[t',f_1',f_2'])$ for $i<j$ and
$(i,j,f_1,f_2) \approx (i,j,f_1',f_2')$, as follows.
\medskip
\newcommand{\RHS}{$\\[.3em] \hspace*{1em}$}
\newcommand{\LHS}{$\\[.4em]$}
\newcommand{\COUNT}{
\refstepcounter{equation}
 \hspace*{\fill}
(\arabic{equation})
}
\noindent
$
\Plowest([t ,\ i,\ j,\  f_1,\  f_2],\ [t',f_1',f_2']) \ =
\COUNT
\label{TAGtreeLOW}
\RHS \Plowest([R_t ,\ i,\ j,\  f_1,\  f_2],\ [t',f_1',f_2'])\ + 
\RHS\deltafunc( (t, f_1, f_2) = (t',f_1',f_2') )\ \cdot
\RHS\hspace{1cm} \Phighest([R_t ,\ i,\ j,\  f_1,\  f_2]);  
\LHS \Plowest([\alpha N , i, j,  -,  -],\ [t,f_1',f_2'])\ =
\COUNT
\label{TAGrecur1LOW}
\RHS \Plowest([\alpha , i, j,  -,  -],\ [t,f_1',f_2'])\  \cdot 
\RHS\hspace{1cm}P([N , j, j,  -,  -])\ + 
\RHS P([\alpha , i, i,  -,  -])\ \cdot 
\RHS\hspace{1cm}\Plowest([N , i, j,  -,  -],\ [t,f_1',f_2']), 
\RHS\mbox{if\ }
\alpha\neq\ep\wedge \neg\domfoot(\alpha N);  
\LHS \Plowest([\alpha N , i, j,  f_1,  f_2],\ [t,f_1',f_2'])\ =
\COUNT
\label{TAGrecur2LOW}
\RHS \deltafunc( f_1=j )  \cdot
\Plowest([\alpha , i, j,  -,  -],\ [t,f_1',f_2'])\ \cdot 
\RHS\hspace{1cm}P([N , j, j,  f_1,  f_2])\ + 
\RHS P([\alpha , i, i,  -,  -])\  \cdot  
\RHS\hspace{1cm}\Plowest([N , i, j,  f_1,  f_2],\ [t,f_1',f_2']), 
\RHS \mbox{if\ }
\alpha\neq\ep\wedge \domfoot(N);  
\LHS \Plowest([\alpha N , i, j,  f_1,  f_2],\ [t,f_1',f_2'])\ =
\COUNT
\label{TAGrecur3LOW}
\RHS\Plowest([\alpha , i, j,  f_1,  f_2],\ [t,f_1',f_2'])\  \cdot 
\RHS\hspace{1cm}P([N , j, j,  -,  -])\ + 
\RHS\deltafunc( i=f_2 )  \cdot \nonumber
        P([\alpha , i, i,  f_1,  f_2])\  \cdot 
\RHS\hspace{1cm}\Plowest([N , i, j,  -,  -],\ [t,f_1',f_2']), 
\RHS\mbox{if\ }
\alpha\neq\ep\wedge \domfoot(\alpha);  
\LHS \Plowest([N, i, j, f_1, f_2],\ [t,f_1',f_2']) \ =
\COUNT
\label{TAGrule1LOW}
\RHS \phi({\bf nil},N)\ \cdot
\RHS\hspace{1cm} \Plowest([\children(N), i, j, f_1, f_2],\ [t,f_1',f_2'])\ + 
\RHS \Plowest([\children(N), i, j, f_1,  f_2],\ [t,f_1',f_2'])\ \cdot
\RHS\hspace{1cm} \sum_{t'\in {\cal A}}
     \phi(t',N)\cdot P([t',  i, j, i,  j])\ + 
\RHS
P([\children(N), f_1, f_2, f_1,  f_2])\ \cdot 
\RHS\hspace{.5cm} {\displaystyle\sum_{t'\in {\cal A}}}
     \phi(t',N)\cdot
\Plowest([t',  i, j, f_1,  f_2],\ [t,f_1',f_2']),
\RHS \mbox{if\ } N\in \myvar\wedge\domfoot(N);  
\LHS \Plowest([N, i, j, -, -],\ [t,f_1',f_2']) \ =
\COUNT
\label{TAGrule2LOW}
\RHS \phi({\bf nil},N)\ \cdot
\RHS\hspace{1cm} \Plowest([\children(N), i, j, -, -],\ [t,f_1',f_2'])\ + 
\RHS  \Plowest([\children(N),  i, j, -, -],\ [t,f_1',f_2'])\ \cdot
\RHS\hspace{1cm} \sum_{t'\in {\cal A}}
     \phi(t',N)\cdot P([t',  i, j, i,  j])\ + 
\RHS {\displaystyle\longsum{\lefteqn{f_1'',f_2''
        (f_1''=  f_2''=i \vee f_1''=  f_2''=j)}\hspace{1.5cm} }}
P([\children(N),  f_1'', f_2'', -, -])\ \cdot 
\RHS\hspace{.5cm} {\displaystyle\sum_{{t'}\in {\cal A}}}
     \phi(t',N)\cdot
\Plowest([t',  i, j, f_1'',  f_2''],\ [t,f_1',f_2']),
\RHS \mbox{if\ }N\in \myvar\wedge  \neg\domfoot(N);  
\LHS \Plowest([a , i, j,  -,  -],\ [t,f_1',f_2']) = 0; 
\COUNT
\label{TAGscanLOW}
\LHS \Plowest([\bot, i, j,  f_1,  f_2],\ [t,f_1',f_2'])  =   0; 
\COUNT
\label{TAGfootLOW}
\LHS \Plowest([\ep, i, j,  -,  -],\ [t,f_1',f_2'])  =   0.
\COUNT
\label{TAGstopLOW}
$

\smallskip
\noindent
The definition of $\Plowest$ parallels the one of $P$ given
in~\S\ref{ss:general}.  In~(\ref{TAGtreeLOW}), the second term in the
right-hand side accounts for the case in which the tree we are
visiting is the ``lowest'' one on which $\Plowest$ should be called.
Note how 
in the above equations 
$\Plowest$ must be called also on nodes that do not dominate
the footnode of the elementary tree they belong to (cf.\ the
definition of $\approx$).  Since no call to $\Phighest$ is possible
through the terms in~(\ref{TAGscanLOW}), (\ref{TAGfootLOW})
and~(\ref{TAGstopLOW}), we must set the right-hand side of these
equations to $0$.

The specification of $\Phighest([\alpha,i,j, f_1, f_2])$ is given
below.  Again, the definition parallels the one of $P$ given
in~\S\ref{ss:general}. \\[-4ex]
\begin{eqnarray}
\label{TAGrecur1HIGH}
\lefteqn{\Phighest([\alpha N, i, j, -, -])\ =}\\&&
\longsum{k(i < k < j)} P([\alpha , i, k,  -,  -])  \cdot
P([N , k, j,  -,  -])\ + \nonumber\\
&&\Phighest([\alpha , i, j,  -,  -])  \cdot
P([N , j, j,  -,  -])\ +\nonumber \\
&&P([\alpha , i, i,  -,  -])  \cdot
\Phighest([N , i, j,  -,  -]), \nonumber \\
&&\mbox{if\ }
\alpha\neq\ep\wedge \neg\domfoot(\alpha N); \nonumber \\
\label{TAGrecur2HIGH}
\lefteqn{\Phighest([\alpha N , i, j,  f_1,  f_2])\ =} \\ &&
\longsum{\hspace{1cm}k(i < k \leq f_1 \wedge k < j)} P([\alpha , i, k,  -,  -])  
\cdot
P([N , k, j,  f_1,  f_2])\ + \nonumber \\
&& \deltafunc( f_1=j )\cdot \Phighest([\alpha , i, j,  -,  -])\ \cdot\nonumber\\
&&\hspace{1cm} P([N , j, j,  f_1,  f_2])\ + \nonumber \\
&&P([\alpha , i, i,  -,  -])  \cdot
\Phighest([N , i, j,  f_1,  f_2]),\nonumber \\
&& \mbox{if\ }
\alpha\neq\ep\wedge \domfoot(N);\nonumber  \\
\label{TAGrecur3HIGH}
\lefteqn{\Phighest([\alpha N , i, j,  f_1,  f_2])\ =} \\ &&
\longsum{\hspace{1cm}k(i<k\wedge f_2\leq k < j)} P([\alpha , i, k,  f_1,  f_2])  
\cdot
P([N , k, j,  -,  -])\ + \nonumber\\
&&\Phighest([\alpha , i, j,  f_1,  f_2])  \cdot
P([N , j, j,  -,  -])\ +\nonumber \\
&&\deltafunc(i=f_2) \cdot P([\alpha , i, i,  f_1,  f_2])  \cdot\nonumber\\
&&\hspace{1cm} \Phighest([N , i, j,  -,  -]), \nonumber \\
&&\mbox{if\ }
\alpha\neq\ep\wedge \domfoot(\alpha); \nonumber \\
\label{TAGrule1HIGH}
\lefteqn{\Phighest([N, i, j, f_1, f_2])
\ =} \\ &&
\phi({\bf nil},N)\cdot 
        \Phighest([\children(N), i, j, f_1, f_2])\ + \nonumber\\
&& \longsum{
\hspace{4cm}
f_1',f_2'
\begin{array}[t]{l}
(i\leq f_1'\leq f_1 \wedge f_2\leq f_2'\leq j\ \wedge \\[0.3ex]
\ (f_1',f_2')\neq (i,j) \wedge (f_1',f_2')\neq (f_1,f_2))
\end{array}
}%
P([\children(N), f_1', f_2', f_1,  f_2])\ \cdot\nonumber  \\
&&\hspace{1cm} \sum_{t\in {\cal A}}
     \phi(t,N)\cdot P([t,  i, j, f_1',  f_2'])\ +\nonumber  \\
&&\Phighest([\children(N), i, j, f_1,  f_2])\ \cdot\nonumber  \\
&&\hspace{1cm} \sum_{t\in {\cal A}}
     \phi(t,N)\cdot P([t,  i, j, i,  j]),\nonumber  \\
&& \mbox{if\ }N\in \myvar\wedge  \domfoot(N);\nonumber  \\
\label{TAGrule2HIGH}
\lefteqn{\Phighest([N, i, j, -, -])
\ =} \\ &&
\phi({\bf nil},N)\cdot 
        \Phighest([\children(N), i, j, -, -])\ + \nonumber\\
&& \longsum{
\hspace{4cm}
f_1',f_2'
\begin{array}[t]{l}
(i\leq f_1' \leq f_2'\leq j\wedge (f_1',f_2')\neq(i,j)\ \wedge \\[0.3ex]
\ \neg( f_1' =f_2'=i\vee f_1' =f_2'=j))
\end{array}
}%
P([\children(N),  f_1', f_2', -, -])\ \cdot\nonumber  \\
&&\hspace{1cm} \sum_{t\in {\cal A}}
     \phi(t,N)\cdot P([t,  i, j, f_1',  f_2'])\ +\nonumber  \\
&&\Phighest([\children(N),  i, j, -, -])\ \cdot\nonumber  \\
&&\hspace{1cm} \sum_{t\in {\cal A}}
     \phi(t,N)\cdot P([t,  i, j, i,  j]),\nonumber  \\
&& \mbox{if\ }N\in \myvar\wedge  \neg\domfoot(N); \nonumber  \\
\label{TAGscanHIGH}
\lefteqn{\Phighest([a , i, j,  -,  -]) = \deltafunc(i+1=j\wedge  a_j=a);} \\
\label{TAGfootHIGH}
\lefteqn{\Phighest([\bot, i, j,  f_1,  f_2]) = 0;} \\
\label{TAGstopHIGH}
\lefteqn{\Phighest([\ep, i, j,  -,  -]) = 0.}
\end{eqnarray}

We can now separate those branches of recursion that terminate on the
given input from the cases of endless recursion.  We assume below that
$\Phighest([R_t, i, j, f_1', f_2']) > 0$.  Even if this is not always
valid, for the purpose of deriving the equations below, this
assumption does not lead to invalid results.  We define a new function
$\Poutside$, which accounts for probabilities of subderivations that
do not derive any words in the prefix, but contribute structurally to
its derivation: \vspace{-1ex}
\begin{eqnarray}
\label{tagoutsideA}
\lefteqn{\Poutside([t, i, j, f_1,  f_2],\ [t',f_1',f_2'])\ = }\\ &&
\frac{\Plowest([t, i, j, f_1,  f_2],\ [t',f_1',f_2'])}%
{\Phighest([R_{t'}, i, j, f_1',  f_2'])};\nonumber \\
\label{tagoutsideB}
\lefteqn{\Poutside([\alpha , i, j, f_1,  f_2],\ [t',f_1',f_2'])\ = }\\ &&
\frac{\Plowest([\alpha , i, j, f_1,  f_2],\ [t',f_1',f_2'])}%
{\Phighest([R_{t'}, i, j, f_1',  f_2'])}.\nonumber
\end{eqnarray}

We can now eliminate the infinite recursion that arises
in~(\ref{tagnontsum1}) and (\ref{tagnontsum2}) by rewriting
$P([t,i,j,f_1,f_2])$ in terms of $\Poutside$:
\begin{eqnarray}
\label{outside1}
\lefteqn{P([t, i, j, f_1,  f_2]) \ =} \\ && 
\longsum{\hspace{2cm}t'\in{\cal A},f_1',f_2'
((i, j, f_1',  f_2')\approx (i, j, f_1, f_2))}%
\Poutside([t, i, j, f_1,  f_2],\ [t',f_1',f_2']) \cdot\nonumber\\
&&\hspace{1cm} \Phighest([R_{t'}, i, j, f_1',  f_2']); \nonumber \\
\label{outside2}
\lefteqn{P([t, i, j, -,  -]) \ =} \\ && 
\longsum{t'\in\{t\}\cup{\cal A},f\in\{-,i,j\}}
\Poutside([t, i, j, -,  -],\ [t',f,f]) \cdot\nonumber\\
&&\hspace{1cm} \Phighest([R_{t'}, i, j, f,  f]). \nonumber
\end{eqnarray}
\noindent
Equations for $\Poutside$ will be derived in the next subsection. 

In summary, terminating computation of prefix probabilities
should be based on equations~(\ref{outside1}) and~(\ref{outside2}), 
which replace~(\ref{TAGtree}), along with 
equations~(\ref{TAGrecur1}) to~(\ref{TAGstop}) and 
all the equations for $\Phighest$.

\subsection{Off-line Equations}
\label{ss:off-line}  

In this section we derive equations for function $\Poutside$
introduced in~\S\ref{ss:on-line} and deal with all remaining cases of
equations that cause infinite recursion. 

In some cases, function $P$ can be computed independently 
of the actual input.  For any $i < n$
we can consistently define the following quantities, 
where $t \in {\cal I} \cup {\cal A}$ and $\alpha \in \myvarbot$ 
or $\children(N)=\alpha\beta$ for some $N$ and $\beta$: 
\vspace{-1.5ex}
\begin{eqnarray*}
H_t & = & P([t, i,i,f,f]); \\
H_\alpha & = & P([\alpha, i,i,f',f']), \\[-4.5ex]
\end{eqnarray*}
where $f=i$ if $t \in {\cal A}$, $f=-$ otherwise, 
and $f'=i$ if $\domfoot(\alpha)$, $f=-$ otherwise.  
Thus, $H_t$ is the probability of all derived trees obtained from $t$, 
with no lexical node at their yields.  Quantities $H_t$ and $H_\alpha$
can be computed by means of a system of equations which can be
directly obtained from equations~(\ref{TAGtree}) to~(\ref{TAGstop}).
Similar quantities as above must be introduced for the case $i=n$. 
For instance, we can set $H'_t = P([t, n,n,f,f])$, $f$ specified as above, 
which gives the probability of all derived trees obtained from $t$
(with no restriction at their yields).

Function $\Poutside$ is also independent of the actual input. 
Let us focus here on the case $f_1, f_2 \not \in \{i,j,-\}$ 
(this enforces $(f_1, f_2) = (f'_1, f'_2)$ below).  
For any $i,j,f_1, f_2 < n$, 
we can consistently define the following quantities.
\vspace{-1.5ex}
\begin{eqnarray*}
L_{t,t'} & = & \Poutside([t, i, j, f_1,  f_2],\ [t',f_1',f_2']); \\
L_{\alpha,t'} & = & \Poutside([\alpha, i, j, f_1,  f_2],\ [t',f_1',f_2']). 
\\[-4.5ex]
\end{eqnarray*}
In the case at hand, $L_{t,t'}$ is the probability of all 
derived trees obtained from $t$ such that 
(i)~no lexical node is found at their yields; and 
(ii)~at some `unfinished' node dominating the foot of $t$, 
the probability of the adjunction 
of $t'$ has already been accounted for, but $t'$ itself
has not been adjoined.  

It is straightforward to establish a system of equations for the
computation of $L_{t,t'}$ and $L_{\alpha,t'}$, 
by rewriting equations~(\ref{TAGtreeLOW}) to~(\ref{TAGstopLOW})
according to~(\ref{tagoutsideA}) and~(\ref{tagoutsideB}).  
For instance,
combining~(\ref{TAGtreeLOW}) and~(\ref{tagoutsideA}) gives
(using the above assumptions on $f_1$ and~$f_2$):
\vspace{-1.5ex}
\begin{eqnarray*}
L_{t,t'} & = & L_{R_t,t'} + \deltafunc(t=t'). \\[-4.5ex]
\end{eqnarray*}
Also, if $\alpha \neq \ep$ and $\domfoot(N)$, 
combining~(\ref{TAGrecur2LOW}) and~(\ref{tagoutsideB}) gives 
(again, using previous assumptions on $f_1$ and~$f_2$;
note that the $H_\alpha$'s are known terms here):
\vspace{-1.5ex}
\begin{eqnarray*}
L_{\alpha N,t'} & = & 
        H_\alpha \cdot L_{N,t'}.  \nonumber \\[-4.5ex]
\end{eqnarray*}
For any $i,f_1, f_2 < n$ and $j = n$, 
we also need to define: 
\vspace{-1.5ex}
\begin{eqnarray*}
L'_{t,t'} & = & \Poutside([t, i, n, f_1,  f_2],\ [t',f_1',f_2']); \\
L'_{\alpha,t'} & = & \Poutside([\alpha, i, n, f_1,  f_2],\ [t',f_1',f_2']). 
\\[-4.5ex]
\end{eqnarray*}
Here $L'_{t,t'}$ is the probability of all 
derived trees obtained from $t$ with a node dominating the 
foot node of $t$, that is an adjunction 
site for $t'$ and is `unfinished' in the same sense as above,
and with lexical nodes only in the portion of the tree 
to the right of that node.   
When we drop our assumption on $f_1$ and $f_2$, 
we must (pre)compute in addition terms of the form 
$\Poutside([t, i, j, i, i],$ $[t',i,i])$ and 
$\Poutside([t, i, j, i, i],$ $[t',j,j])$ for $i < j < n$, 
$\Poutside([t, i, n, f_1, n],$ $[t',f'_1,f'_2])$ for $i<f_1 < n$,
$\Poutside([t, i, n, n, n],$ $[t',f'_1,f'_2])$ for $i < n$,
and similar. 
Again, these are independent of the choice of $i$, $j$ and $f_1$.
Full treatment is omitted due to length restrictions.

\section{Complexity and concluding remarks}
\label{s:complexity}

We have presented a method for the computation of the prefix
probability when the underlying model is a Tree Adjoining Grammar.
Function $\Phighest$ is the core of the method. Its equations can be
directly translated into an effective algorithm, using standard
functional memoization or other tabular techniques.  It is easy to see
that such an algorithm can be made to run in time $\order{n^6}$, where
$n$ is the length of the input prefix.

All the quantities introduced in~\S\ref{ss:off-line} ($H_t$,
$L_{t,t'}$, etc.) are independent of the input and should be computed
off-line, using the system of equations that can be derived as
indicated.  For quantities $H_t$ we have a non-linear system, since
equations (2) to (6) contain quadratic terms.  Solutions can then be
approximated to any degree of precision using standard iterative
methods, as for instance those exploited in~\cite{Stolcke}.  Under the
hypothesis that the grammar is consistent, that is $\Pr(L(G)) = 1$,
all quantities $H'_t$ and $H'_\alpha$ evaluate to one.  For quantities
$L_{t,t'}$ and the like,~\S\ref{ss:off-line} provides linear systems
whose solutions can easily be obtained using standard methods.  Note
also that quantities $L_{\alpha,t'}$ are only used in the off-line
computation of quantities $L_{t,t'}$, they do not need to be stored
for the computation of prefix probabilities (compare equations for
$L_{t,t'}$ with~(\ref{outside1}) and~(\ref{outside2})).

We can easily develop implementations of our method that can compute
prefix probabilities incrementally.  That is, after we have computed
the prefix probability for a prefix $a_1 \cdots a_n$, on input $a_{n+1}$
we can extend the calculation to prefix $a_1 \cdots a_n a_{n+1}$
without having to recompute all intermediate steps that do not depend
on $a_{n+1}$. This step takes time $\order{n^5}$.

In this paper we have assumed that the parameters of the stochastic
TAG have been previously estimated. In practice, smoothing to avoid
sparse data problems plays an important role. Smoothing can be handled
for prefix probability computation in the following ways.  Discounting
methods for smoothing simply produce a modified STAG model which is
then treated as input to the prefix probability computation.
Smoothing using methods such as deleted interpolation which combine
class-based models with word-based models to avoid sparse data
problems have to be handled by a cognate interpolation of prefix
probability models.

{\footnotesize 
\newcommand{\noop}[1]{}\newcommand{\id}[1]{#1}

}

\end{document}